\documentclass[letterpaper, 10 pt, conference]{ieeeconf}  

\IEEEoverridecommandlockouts                            
\usepackage{ragged2e}
\usepackage{cite}
\usepackage{ulem}
\usepackage{amsfonts}
\usepackage{graphicx}
\usepackage{amsmath}
\usepackage{booktabs}
\usepackage{multirow}
\usepackage{hyperref}
\usepackage[utf8]{inputenc}
\usepackage[T1]{fontenc}
\usepackage{caption}
\usepackage{xcolor}
\usepackage{fancyhdr}
\hypersetup{hidelinks}

\newcommand\blfootnote[1]{%
  \begingroup
  \renewcommand\thefootnote{}\footnote{#1}%
  \addtocounter{footnote}{-1}%
  \endgroup
}

\title{\LARGE \bf
Robust Quadrupedal Locomotion via Risk-Averse Policy Learning \vspace{-1em}
}

\author{Jiyuan Shi$^{1,2 \,*}$, Chenjia Bai$^{2 \,\dag}$, Haoran He$^{2,3}$, Lei Han$^{4}$, Dong Wang$^{2}$, Bin Zhao$^{2,5}$,\\ Mingguo Zhao$^{1}$, Xiu Li$^{1}$, Xuelong Li$^{2,5}$ 
\thanks{$^{1}$Tsinghua University, China.$^{2}$Shanghai Artificial Intelligence Laboratory, China.$^{3}$Shanghai Jiao Tong University, China.$^{4}$Tencent Robotics X, China.$^{5}$Northwestern Polytechnical University, China.$^{\dag}$Corresponding author: Chenjia Bai.}
\thanks{$^{*}$This work was conducted during the internship of Jiyuan Shi at Shanghai Artificial Intelligence Laboratory.}
}

\begin{document}

\maketitle
\thispagestyle{empty}
\pagestyle{empty}

\begin{abstract}

The robustness of legged locomotion is crucial for quadrupedal robots in challenging terrains. Recently, Reinforcement Learning (RL) has shown promising results in legged locomotion and various methods try to integrate privileged distillation, scene modeling, and external sensors to improve the generalization and robustness of locomotion policies. However, these methods are hard to handle uncertain scenarios such as abrupt terrain changes or unexpected external forces. In this paper, we consider a novel risk-sensitive perspective to enhance the robustness of legged locomotion. Specifically, we employ a distributional value function learned by quantile regression to model the aleatoric uncertainty of environments, and perform risk-averse policy learning by optimizing the worst-case scenarios via a risk distortion measure. Extensive experiments in both simulation environments and a real Aliengo robot demonstrate that our method is efficient in handling various external disturbances, and the resulting policy exhibits improved robustness in harsh and uncertain situations in legged locomotion. Videos are available at \href{https://risk-averse-locomotion.github.io/}{https://risk-averse-locomotion.github.io/}.
\end{abstract}
\blfootnote{This work has been submitted to the IEEE for possible publication. Copyright may be transferred without notice, after which this version may no longer be accessible.}
\section{Introduction}

Quadrupedal robots are widely recognized for their exceptional agility and remarkable capability to traverse complex terrains, which is crucial for scenarios such as industrial inspections and firefighting. Previous methods adopt Model Predictive Control (MPC) for quadrupedal robots, while it typically requires precise dynamics modeling with domain-specific knowledge \cite{winkler2018gait,kim2019highly} and there is a trade-off between the model accuracy and computational complexity. Recently, model-free Reinforcement Learning (RL) demonstrated impressive performance in legged locomotion without dynamics modeling \cite{SR-2019}. The RL policy can be trained by interacting with simulated environments, especially the parallel simulator like Isaac Gym \cite{isaac-gym}, allowing robots to traverse various complex terrains such as rocks, stairs, snow, and beaches \cite{Isaac-Gym-walk}. 

Existing RL-based approaches try to enhance the robustness and generalization ability of locomotion policies via privileged distillation, scene modeling, and external sensors. Specifically, privileged distillation methods adopt a teacher-student architecture to help the student policy infer the privileged state of the true environment \cite{SR-2020,RMA-2021,DreamWaQ,margolis2022rapid}; scene modeling methods explicitly learn the scene geometry \cite{VolumetricMemory2023}, terrain traversability \cite{Traversability-2022}, or via volumetric models; other methods equipped robots with cameras or LiDAR to enhance their terrain traversal capabilities \cite{SR-2022-wild,Egocentric-2022}. Although these methods show extraordinary performance in legged locomotion, they are still hard to handle abrupt events in locomotion such as unexpected terrain changes or external forces. The reason is they only infer the information of the current state from historical or privileged information, without considering the possible risk events in the future. Meanwhile, external sensors are often unreliable and need additional models for understanding and reasoning in the environment. In addition, RL-based methods often adopt domain randomization \cite{haarnojaLearningWalkDeep2019,imaiVisionGuidedQuadrupedalLocomotion2022} to enhance the robustness in sim-to-real transfer, while excessive randomization may lead to an overly conservative policy.

\begin{figure}[t]
\centering
\includegraphics[width=0.49\textwidth]{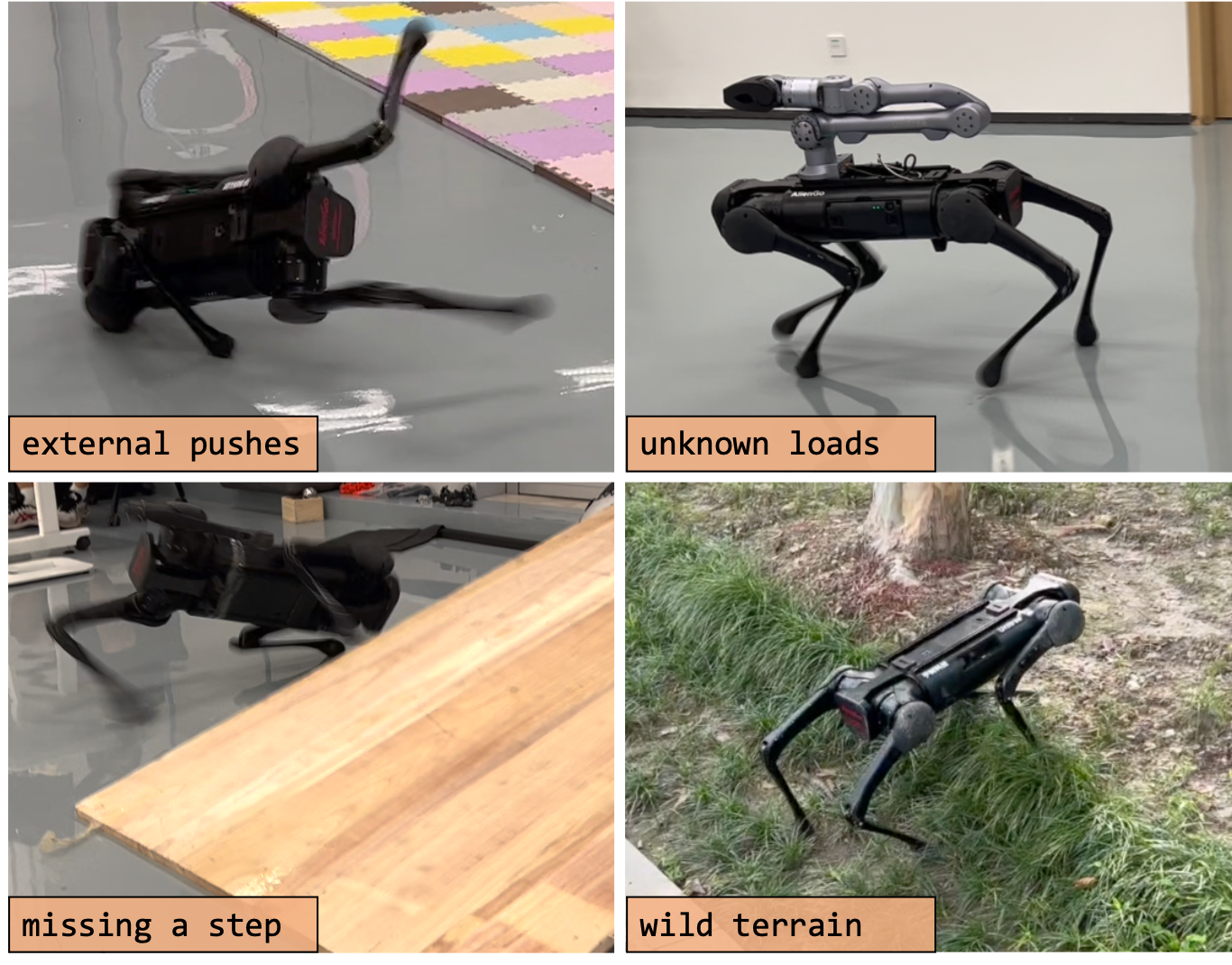}
\caption{We trained a robust locomotion controller through risk-sensitive RL. The robot demonstrated robustness when encountering risks in the environment, such as sudden pushes and missing a step.}
\vspace{-1em}
\vspace{-0.4cm}
\label{fig:fig1}
\end{figure}

Different from the above methods, we explicitly consider modeling the risks in legged locomotion and take a \emph{risk-sensitive} perspective in policy learning to enhance the robustness of locomotion policy. Specifically, optimizing the expected return like previous methods cannot avoid risky events (e.g., falling) since avoiding risks may slow down the robot's speed and reduces the expected return \cite{rudinLearningWalkMinutes2022}. However, we prioritize preventing the robot from falling which may cause hardware damage rather than precisely tracking the speed and directional commands. Such a risk-averse perspective can enhance the robot's ability to withstand uncertain disturbances \cite{shen2014risk}. 

In this work, we propose Risk-Averse Legged Locomotion (RALL), which performs risk-averse policy learning via a distributional value function to estimate the aleatoric uncertainty (i.e., risks) \cite{decomposition-2018} of the environment. The value distribution is learned by quantile regression in an actor-critic framework, then the agent can obtain risk-sensitive policies by optimizing the varying level of risks. To obtain a robust policy in legged locomotion, we optimize the bottom percentile of value distribution to learn a policy that performs well in the worst case via Conditional Value-at-Risk (CVaR), resulting in a \emph{risk-averse} policy. Further, considering environments which contain multiple types of terrains, the robot may need to switch between policies with different risk preferences according to the specific circumstance. To this end, we propose IQR (Interquartile Range) of the quantile function as a risk-level measurement of the environment. The agent can choose to use a risk-averse policy or an ordinary policy based on IQR. For implementation, we design several proprioception-based risks in legged locomotion where each kind of risk occurs following a Bernoulli distribution. The resulting controller is able to generalize well across diverse perturbations, transcending the limitations imposed by the training environment's diversity. 

To the best of our knowledge, RALL provides the first risk-sensitive policy learner in locomotion control of quadruped robots. Through a comprehensive evaluation in both simulation and real-world experiments, we demonstrate that RALL significantly enhances the robustness of locomotion. We show that RALL equips a real Unitree Aliengo\cite{AliengoMultifunctionalIndustrial} robot with the capability of traversing challenging terrains, withstanding dynamic loads, and resisting substantial external disturbances, without relying on the external sensors and extensive randomization. The main contributions of this paper are as follows:
\begin{itemize}
\item We present a novel perspective on achieving a robust locomotion controller via distributional value function and risk-sensitive policy learning. The resulted controller enabled the robot to resist heavy impact and traverse challenging terrains.
\item We propose IQR as a risk measure in legged locomotion to enable the robot to choose policy with different risk preferences according to the current environment. IQR can be combined with other locomotion policies to enhance their robustness by switching to a risk-averse policy for environments with a higher IQR. 
\item The simulated and real-world experiments on Aliengo robot show that the RALL agent performs robustness and can traverse challenging terrains, endure dynamic loads, and recover from significant external pushes.
\end{itemize}

\section{Related Work}

\subsection{RL-based Quadruped Locomotion}

RL-based methods for locomotion control in quadrupedal robots demonstrate the capability to traverse complex terrains \cite{margolis2022rapid,Egocentric-2022,choiLearningQuadrupedalLocomotion2023}. However, the robustness of quadruped robot is still an open problem \cite{wellhausenArtPlannerRobustLegged2023,jiConcurrentTrainingControl2022}. A prevalent approach to bridge the reality gap and enhance robustness is privileged distillation, which involves a teacher policy to encode privileged information (e.g., elevation map of the surrounding terrain and randomized parameters) into a latent vector \cite{SR-2020,SR-2022-wild,RMA-2021,margolis2022rapid,Egocentric-2022,he2023privileged, VolumetricMemory2023}. The student is trained via supervised learning using accessible states on real robots. However, this framework is inefficient in training independent policies \cite{jiConcurrentTrainingControl2022} and cannot predict risky events in the future. DreamWaQ \cite{DreamWaQ} leveraged an asymmetric actor-critic architecture and a context-aided estimator network to infer the terrain properties, while it still needs privileged information in training. In contrast, we only use proprioceptive feedback in policy training without privileged knowledge or teacher-student architectures. Other methods also incorporate external sensors like cameras or LiDAR in the RL framework for scene understanding \cite{vision-lclr}, while it relies on vision models and additional computing. In RALL, we find the risk-averse policy can perform robustly without external sensors, and our method can be easily combined with vision-based policies \cite{Egocentric-2022}. Several works propose learning a recovery controller to enable robots to recover from a fall \cite{lee2019robust, maLearningArmAssistedFall2023}. Nevertheless, our work has a different objective to prevent the agent from falling since it can easily result in hardware damage. 

Beyond robustness, there exist several other directions to enhance the abilities of quadruped robots. 
Researchers have enriched the ability of quadruped robots by learning diverse and agile locomotion skills like trotting, pronking, and jumping via developing different reward functions \cite{walk-these-ways}, performing imitation learning from animals \cite{imitate-animal} or reference trajectories \cite{li2023learning}, or transferring experience from existing controllers \cite{smithLearningAdaptingAgile2023}. Especially, the quadruped robot can learn sidestep, dive, and jump skills to become a qualified goalkeeper \cite{gokeeper}, or fast kicking skill to shoot a ball \cite{shoot}. Other works also improve the sample efficiency \cite{dayDreamer,20minutes} and generalization ability via developing unified policy for quadrupedal robots with similar morphologies \cite{GenLoco}, performing whole-body control of manipulation and locomotion \cite{whole-body}, and combining navigation with legged locomotion \cite{yokoyama2023adaptive,fu2022coupling}.

\subsection{Distributional RL and Risk-Sensitive Learning}

The distributional perspective in RL has a rich history\cite{sobel1982variance,jaquette1973markov,white1988mean}. In deep RL, C51 \cite{C51} first applies distributional Bellman equation to learn the value distribution, whose support is a set of atoms. In distributional RL, the value function estimates the whole distribution of return rather than its mean. An improvement over C51 is QR-DQN \cite{dabneyQRDQNDistributionalReinforcement2017}, which parameterizes the values of fixed quantiles and minimizes the Wasserstein distance to a target distribution. IQN \cite{dabneyIQNImplicitQuantile2018} further extends the discrete
quantile fractions to a continuous function by using a distortion function. IQN can approximate different quantile of return distribution and learn policies under different risk measures, such as Wang \cite{muller1997integral}, CPT \cite{prashanth2016cumulative}, and Conditional Value at Risk (CVaR) \cite{rockafellar2000optimization,chow2014algorithms}, leading to risk-averse or risk-seeking policies \cite{urpiRiskAverseOfflineReinforcement2021,jiangLearningDiverseRisk2023,baiMonotonicQuantileNetwork2022}. The value distribution can  reflect the aleatoric uncertainty of the environment \cite{baiMonotonicQuantileNetwork2022}, which is used for risk-sensitive learning in online exploration \cite{bharadhwaj2020conservative,mavrin2019distributional,mavor2022stay,tang2019worst} to avoid risk events or in an offline setting to learn risk-averse policy \cite{ORAAC, CODAC}. 

While there have been numerous studies, the application of return distribution in real quadruped robots is limited. Haarnoja et al. use a C51-style agent \cite{C51} for controlling a biped robot playing soccer \cite{haarnojaLearningAgileSoccer2023}, without considering risk-sensitive learning. To the best of our knowledge, RALL proposes the first risk-sensitive locomotion policy via distributional value function. In addition, we use IQR of quantile distribution to represent the risk level of the environment and switch policy based on the risk level.

\section{METHOD}
\subsection{Problem Definition}
We model the problem of robust locomotion control as a Markov Decision Process (MDP) as $\mathcal M=(\mathcal S, \mathcal A, \mathcal P, R, \gamma)$, where $\mathcal S $ is the state space, $\mathcal A $ is the action space, $\mathcal P(\cdot|s, a) $ is the transition probability, $R: \mathcal S \times \mathcal A \rightarrow \mathcal R$ is the stochastic reward function, and $\gamma \in [0,1)$ is the discount factor. Since we do not use the privileged information in training, the observation $\mathbf{o}_t$ of our framework is equivalent to the state $\mathbf{s}_t$, then we have $\mathbf{s}_t = \mathbf{o}_t$ and 
\vspace{-0.15cm}
\begin{equation}
\mathbf{o}_t = [\mathbf v_t, \boldsymbol{\omega}_t, \mathbf g_t, \mathbf c_t, \mathbf q_t, \mathbf{\dot q}_t, \mathbf{a}_{t-1}] \in \mathbb R^{48},
\vspace{-0.2cm}
\end{equation}
where $\mathbf v_t$ is the base linear velocity, $\boldsymbol{\omega}_t$ is the base angular velocity, $\mathbf g_t$ is the gravity vector in the body frame, $\mathbf c_t\in \mathbb R^3$ is the velocity command, $\mathbf q_t$ is the joint angle, $\mathbf{\dot q}_t$ is the joint velocity, and $\mathbf{a}_{t-1}$ is the last action. We use the target joint position as action, and the low-level torque command is calculated via a PD controller.

Although it's hard to anticipate risks in the environment, the proprioception of the robot could provide clues to indicate whether the robot is facing potential risks. For instance, when the robot's roll angle exceeds a certain threshold, the robot is prone to roll over. In RALL, we introduce risks by adding penalization terms to the reward function. To be specific, when certain state components of the robot exceed predefined thresholds, we apply a relatively large penalization to the reward with a probability of $p$, so the final reward function can be presented as
\vspace{-0.15cm}
\begin{equation}
\label{eq:risk}
    \boldsymbol{r} = \boldsymbol r_{\text{task}} - \sum_{i=1}^{M}w_i\mathbb I_{|s_i|>\bar s_i} \cdot \mathcal B_{p} ,
\vspace{-0.2cm}
\end{equation}
where $\boldsymbol r_{\text{task}}$ is the reward function to accomplish the locomotion task. The details of $\boldsymbol r_{\text{task}}$ are given in Table \ref{tab:reward}. The penalty term in~\eqref{eq:risk} indicates risks, where $M$ is the number of risks, $w_i$ is the weight of each kind of risk, $\mathbb I $ is an indicator function, $s_i$ and $\bar s_i$ are the state and its risk threshold, and $\mathcal B_{p}$ is a Bernoulli variable with a probability of $p$. Table \ref{tab:risk} gives the detailed setup of risks. 

In~\eqref{eq:risk}, we adopt a very small probability (i.e., $p=10^{-4}$) for risk events. Thus, the risk penalty can barely affect the expected return but will have a significant impact on the value distribution, especially in the worst case.

\subsection{Distributional Value Function}
\label{secIII-B}

To estimate the return distribution for risk-sensitive learning, we propose to train a distributional critic based on the risk-injected reward function in~\eqref{eq:risk}. The learned value distribution will be further used to 1) obtain a risk-averse policy and 2) estimate the current risk level. 

Normally, the objective of RL is to maximize the expected cumulative return $\mathbb{E}[Z^\pi(s,a)]$, where $Z^\pi(s, a)$ is the return distribution. We have $Z^\pi(s,a)=\Sigma^\infty_{t=0} \gamma^t R(s_t,a_t)$ is a random variable representing the sum of discounted rewards for the agent following policy $\pi$.
Many standard RL methods estimate the action-value function $Q^\pi(s,a)=\mathbb{E}[Z^\pi(s,a)]$, which could be characterized by the Bellman equation $Q^\pi(s,a)=\mathbb{E}[R(s,a)] + \gamma \mathbb{E}_{\mathcal P, \pi} [Q^\pi(s',a')]$. In distributional RL, the action-value distribution can be learned using distributional Bellman operator~\cite{C51}, as
\vspace{-0.15cm}
\begin{equation}
\mathcal T^\pi Z(s,a): \overset{D}{=} R(s,a) + \gamma Z(S',A'),
\label{eq:4}
\vspace{-0.15cm}
\end{equation}
where $S' \sim \mathcal P(\cdot|s,a), A' \sim \pi(\cdot|s')$, and $Y \overset{D}{:=} U$ denotes that two random
variables have equal probability laws. In the following, we denote $F_Z(z) =
Pr(Z\leq z)$ as the cumulative
density function (CDF) of the distribution $Z$, and $F^{-1}_Z(\tau)$ as the quantile function (i.e., inverse CDF). For $\tau \in [0,1]$, $F^{-1}_Z(\tau) := \inf \{y \in \mathbb R : \tau \leq F_Y(y)\}$. Theoretical work \cite{C51} shows the distributional Bellman operator is a contraction in the $p$-Wasserstein metric, which measures the optimal transport between distributions, as
\vspace{-0.4cm}
\begin{equation}
W_p(Z, \mathcal T^{\pi} Z)=\left(\int_0^1\Big|F_{Z}^{-1}(\omega)-F_{\mathcal T^{\pi} Z}^{-1}(\omega)\Big|^p d \omega\right)^{1/p}.
\end{equation}

\captionsetup[table]{justification=raggedright, singlelinecheck=false}
\begin{table}[t]
\centering
\caption{Definition of reward functions. Here $\tau$ refers to joint torque.} 

\label{tab:reward}
\fontsize{10}{12}\selectfont
\begin{tabular}{lll}
\toprule
Reward Term & Definition & Weight \\
\hline
 linear velocity tracking &     $e^{-(\mathbf v^{\text{cmd}}_{xy} - \mathbf v_{xy})^2/\sigma}$ & 5 \\
 angular velocity tracking&     $e^{-(\boldsymbol{\omega}^{\text{cmd}}_{\text{yaw}} -\boldsymbol{\omega}_{\text{yaw}})^2/\sigma}$ & 0.5 \\
 linear velocity penalty       &     $v_z^2$&   -1.0\\
 angular velocity penalty &  $\boldsymbol{\omega}_{\text{roll,pitch}}^2$ &  -0.05\\
 joint acceleration       &     $\ddot{\mathbf q}^2$  &  -2.5e-7  \\
 torques                  &     $\mathbf \tau^2$&  -2e-5 \\
 action magnitude &  $\mathbf a^2$  & -0.01 \\
 collision &   $n_\text{collision}$&  -1e-3\\
 action rate &  $(\mathbf a_t - \mathbf a_{t-1})^2$ &   -0.01 \\
 torque smooth &  $(\mathbf \tau_t - \mathbf \tau_{t-1})^2$ & -3e-4 \\
 feet air time &  $\Sigma_{f=0}^4(\mathbf t_{air,f}-0.5)$ & 2 \\
\bottomrule
\end{tabular}
\end{table}

\begin{table}[t]
\vspace{-0.2cm}
\centering
\caption{Definition of risks. The risk terms will be added to the reward function defined in~\eqref{eq:risk}.}
\label{tab:risk}
\fontsize{10}{12}\selectfont
\begin{tabular}{lll}
\toprule
Risk Term & Threshold & Weight \\
\hline
base pitch     &  0.5 rad &  20\\
base roll      &  1 rad &  100\\ 
joint velocity  &  10 rad$\cdot \text{s}^\text{-1}$ & 100 \\
joint acceleration  &  1000 $\text{rad}\cdot \text{s}^{\text{-2}} $ & 100 \\
joint torque    &  40 N·m  & 150 \\ 
\bottomrule
\end{tabular}
\vspace{-0.3cm}
\end{table}

\vspace{-0.3cm}
In order to take risk into account, we follow Implicit Quantile Network (IQN) \cite{dabneyIQNImplicitQuantile2018} to estimate the quantile function by using a continuous quantile function $Z_\tau(s,a;\theta):=F^{-1}_{Z(s,a)}(\tau)$ parameterized by $\theta$. For two samples $\tau,\tau' \sim U([0,1])$, the temporal difference (TD) error is
\begin{equation}
    \delta_{\tau,\tau'} = r + \gamma Z_{\tau'}(s',a';\theta)-Z_\tau(s,a;\theta),
\end{equation}
The overall critic loss is given by
\begin{equation}
    \mathcal L_{\text{critic}}(\theta) = \frac{1}{N'}\sum^N_{i=1}\sum^{N'}_{j=1} \rho_{\tau}^\kappa(\delta_{\tau_i,\tau_j}),
\end{equation}
where $N$ and $N'$ are the number of samples , and $\rho_{\tau_i}^\kappa$ is the quantile Huber loss \cite{dabneyQRDQNDistributionalReinforcement2017} defined by
\begin{align}
\begin{split}
&\rho_\tau^\kappa(\delta_{\tau_i,\tau_j}) = \lvert \tau - \mathbb I_{\{\delta_{\tau_i,\tau_j}<0 \}} \rvert \mathcal L_\kappa(\delta_{\tau_i,\tau_j}),  \; \text{where} \\
    &\mathcal{L}_\kappa(\delta_{\tau_i,\tau_j})=\left\{\begin{array}{ll}
\frac{1}{2} \delta_{\tau_i,\tau_j}^2, & \text { if }|\delta_{\tau_i,\tau_j}| \leq \kappa \\
\kappa\left(|\delta_{\tau_i,\tau_j}|-\frac{1}{2} \kappa\right), & \text { otherwise }
\end{array} ,\right.
\end{split}
\end{align}

In continuous action settings, the next state-action value distribution $Z(S',A')$ in (\ref{eq:4}) is a mixture distribution of all possible state-action value distributions, which is infeasible to compute. Thus, we follow \cite{nam2021gmac} to avoid this issue by directly approximating the next state value distribution $Z(S')$ instead of $Z(S',A')$, where $Z(S')$ can be integrated into policy gradient algorithms.

\subsection{Risk-Sensitive Policy Learning}

\begin{figure}[t]
\centering
  \includegraphics[width=0.48\textwidth]{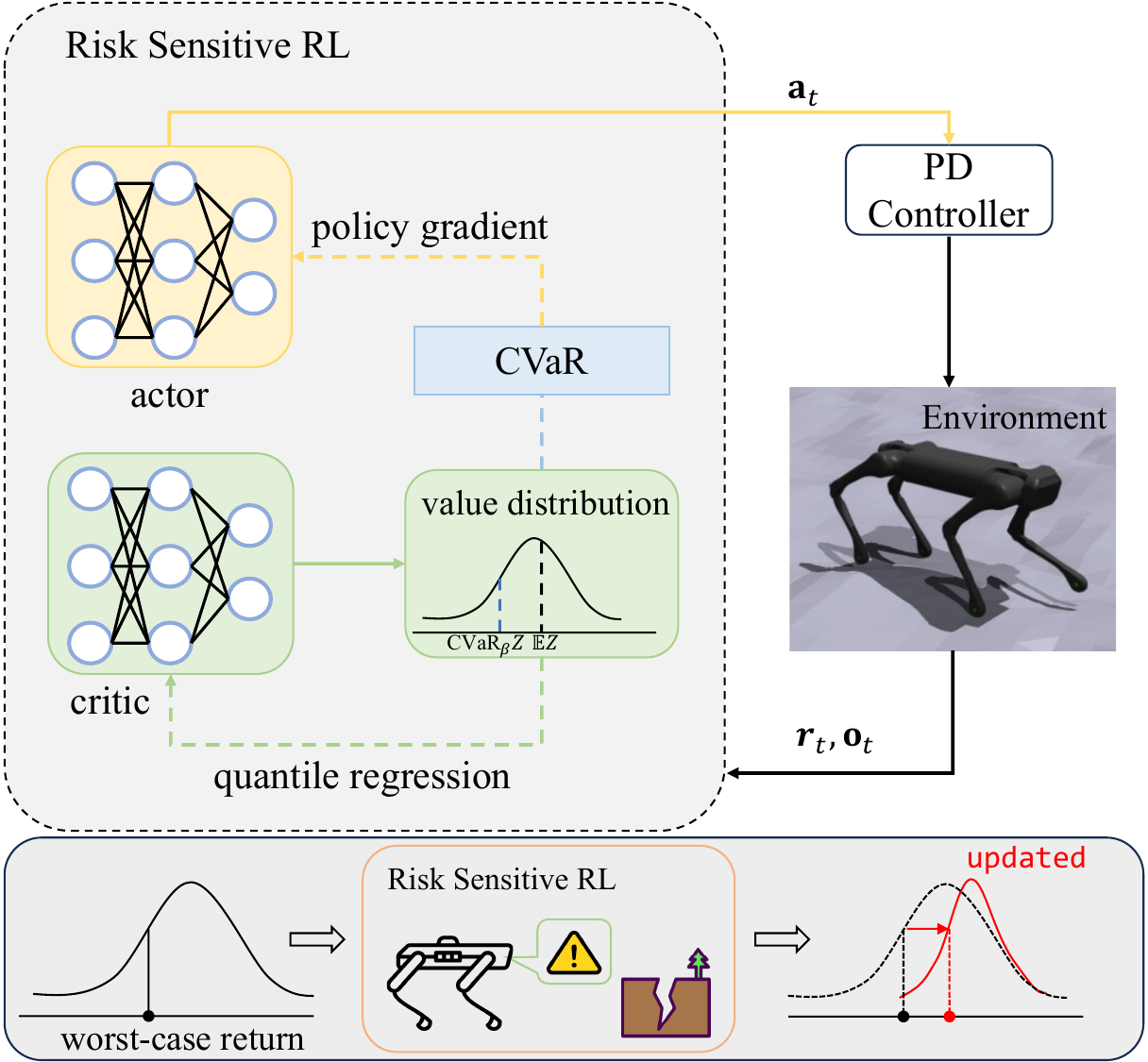}
  \caption{Overall framework of our method. The critic network estimates the value distribution, and the risk-averse policy is obtained by optimizing the CVaR objective. The policy is supposed to perform well under worst-case scenarios.}
  \vspace{-0.4cm}
  \label{fig:framework}
\end{figure}

Based on the distributional value distribution, we adopt the policy gradient method to learn the policy, as
\begin{equation}
    \nabla_{\phi} J(\phi) = \mathbb E\left[ \sum_{t=0}^{T} A_t \nabla_{\phi} \log \big(\pi_{\phi}(a_t\mid s_t)\big)\right],
\end{equation}
where $A_t$ is the advantage function of the value distribution, and $\phi$ is the parameter of the actor-network. According to discussions in Section \ref{secIII-B}, $A_t$ is given by $A_t = r_t + \gamma V(s_{t+1}) - V(s_t)$, where $V(s) := \mathbb{E} \left[Z_\tau(s)\right]$. 

In RALL for legged locomotion, we aim to learn policies with specific \emph{risk preferences} by applying a distortion function \cite{urpiRiskAverseOfflineReinforcement2021} to $\tau$. Let $\beta: [0,1] \rightarrow [0,1] $ be a distortion risk measure, then the distorted expectation of $Z$ under $\beta$ is
\begin{equation}
    V_{\beta}(s) := {\mathbb E}_{\tau \sim U([0,1])} \left[Z_{\beta(\tau)}(s) \right].
\end{equation}
The goal of RALL is to find a risk-sensitive policy by maximizing the distorted expectation of $Z$. A commonly used risk-aversion measure is conditional value-at-risk (CVaR) as $
\text{CVaR}(\eta, \tau) = \eta \tau$, which simply changes the sampling distribution from $\tau \sim U([0,1])$ to $\tau \sim U([0,\eta])$. Then the CVaR objective of the policy is 
\begin{equation}
    \text{CVaR}_\eta (Z(s)) = \mathbb E\left[Z_{\tau \sim U([0,\eta])}(s) \right],
\end{equation}
and the policy gradient under the CVaR measure is given as,
\begin{align}\label{eq:loss=actor}
\begin{split}
    \nabla_{\phi} J(\phi) &= \mathbb E\left[ \sum_{t=0}^{T} A_t^{\text{CVaR}} \nabla_{\phi} \log (\pi_{\phi}(a_t\mid s_t))\right], \text{where}\\
    A_t^{\text{CVaR}} &= r_t + \gamma  \text{CVaR}_\eta (Z(s_{t+1})) - \text{CVaR}_\eta (Z(s_t)).
\end{split}
\end{align}
Intuitively, since $\text{CVaR}_\eta$ only considers the bottom quantiles are less than $\eta$ (with $\eta<1$), the policy is learned to be risk-averse and performs well in worst-case scenarios \cite{urpiRiskAverseOfflineReinforcement2021}. 

In quadrupedal locomotion, we consider risk events such as sudden changes in terrain, external disturbance, and uncertain load. Such scenarios may bring large penalties in reward function, which can be learned by bottom quantile functions. By performing policy gradient in~\eqref{eq:loss=actor}, we obtain a robust policy to exhibit favorable performance in worst-case scenarios, thereby enhancing the robot's resistance to disturbances. We set $\eta$ to be 0.5 in practice. The schematic diagram of our approach is illustrated in Fig.~\ref{fig:framework}. We also use advanced update tricks in policy gradients such as GAE\cite{schulman2015high} and PPO-Clip\cite{schulman2017proximal} to improve the performance. For real-world applications, we conduct domain randomization in simulation to facilitate sim-to-real transfer.

\subsection{Risk-Aware Meta Controller}
\label{sec:method-c}

In RALL, one consideration is that the CVaR objective only optimizes the worst-case scenario, which may lead to suboptimal performance in normal situations. 

An intuitive solution would be switching control policies according to specific circumstances. Specifically, for relatively simple environments, we can employ a risk-neutral policy by setting $\eta=1$ in the above objectives; while for scenarios with increased risk events, the controller should switch to a conservative policy obtained via the optimization of $\text{CVaR}_{0.5}$. 

Since we focus on the robust locomotion control of quadrupedal robots without external sensors, devices such as camera or LiDAR is infeasible in estimating the surrounding information. Fortunately, given the value distribution, we can use the return distribution as a metric for assessing the risk level of the current environment, which has been verified in previous  distributional RL works \cite{mavor2022stay,mavrin2019distributional}. 
In RALL, we propose using the variance of the quantile distribution as the risk measure. Instead of directly calculating the variance, we employ the \emph{interquartile range (IQR)} to estimate the aleatoric uncertainty of returns, which is given by
\begin{equation}
    IQR = Q_3 - Q_1, \quad Q_3 = F^{-1}_Z\left(0.75\right), Q_1 = F^{-1}_Z\left(0.25\right).
\end{equation}
Compared to variance, the advantage of using IQR is that it is less affected by outliers, which makes it suitable for locomotion tasks in quadruped robots.

\section{Experiments}

\subsection{Simulation}
We train our policy in Isaac Gym simulation \cite{isaac-gym} based on the open-source framework in \cite{Isaac-Gym-walk}. The actor and critic networks have the same hidden dimensions of $[512, 256, 128]$. The critic network outputs estimated values of 64 quantiles, which are sampled from $U([0,1])$, and optimization objective of the policy is distorted by $\text{CVaR}_{0.5}$. For the actor-critic algorithm, we set the clipping range, generalized advantage estimation factor $\lambda$, and discount factor $\gamma$ to 0.2, 0.95, and 0.99, respectively, with a learning rate of 1e-3. In order to facilitate the sim-to-real transfer, we incorporate domain randomization throughout the training process, as detailed in Table \ref{tab:dr}. We introduce Gaussian noise to state to make the robot robust against observation errors. The robot is trained on various terrains in simulation, including smooth slopes, rough slopes, and discrete obstacles. We employ the terrain curriculum introduced by \cite{Isaac-Gym-walk} to enable the robot to traverse challenging terrains progressively. 

\begin{table}[b]
\vspace{-0.4cm}
\centering
\caption{Domain randomization terms and their ranges. We simulate pushing on the robot by randomly introducing velocity perturbations to its base. The push interval is 5s.}
\label{tab:dr}
\fontsize{10}{12}\selectfont
\begin{tabular}{lll}
\toprule
Randomized Term & Range & Unit \\
\hline
friction coefficient     &  [0.5,1.5] &  -\\
base mass      &  [-0.5, 1.0] &  kg \\ 
push           &  [-1,1] & m/s \\
\bottomrule
\end{tabular}
\end{table}

We train 4096 agents parallelly on a PC with a 32-core Intel i9-13900K CPU @ 5.5GHz, 128GB RAM, and an NVIDIA RTX 4090 GPU. We train the policy for 6000 iterations, which take approximately 7 hours.

\subsection{Hardware Setup}

We use Unitree Aliengo \cite{AliengoMultifunctionalIndustrial} robot for real-world experiments. The robot has 12 degrees of freedom and weighs about 21kg. The computations are performed on an onboard NVIDIA Jetson TX2. The policy runs at 50Hz and the target joint angles were tracked by a PD controller at a frequency of 200 Hz. The PD gains are $K_p=50$ and $K_d=0.8$, respectively. During the evaluation, we send linear and angular velocity commands to the robot from a remote host. The command was updated at a frequency of 50Hz.

\captionsetup[table]{justification=justified} 
\begin{table}[t]
\centering
\caption{Robustness evaluation for robot carrying challenging payloads. The masses of the ball, container, and robot arm are about 2kg, 0.1kg, and 4.5kg, respectively. For experiments with external pushes, we randomly applied a force of 100N in the x/y direction on the robot's body every 300 steps, and this force was sustained for 50 steps.}
\label{tab:result1}
\fontsize{10}{12}\selectfont
\begin{tabular}{ll|cccc}
\toprule
\multicolumn{2}{c|}{Experiment setup}  & \multicolumn{4}{c}{Time to Fall (TTF)}\\  \hline
\multirow{2}{*}{Load type}  & Terrain                 & \multicolumn{2}{c}{Flat}      & \multicolumn{2}{c}{Random}    \\ \cline{2-6} 
                            & Push              & False & True & False & True                \\ \hline
\multirow{4}{*}{Ball}       & Baseline                     & 0.539            & 0.212       & 0.292            & 0.189       \\
                            & DR & 0.233            & 0.165       & 0.205            & 0.154       \\
                            & RMA                          & 0.338            & 0.224       & 0.305            & 0.189       \\
                            & RALL                   & \textbf{0.963}            & \textbf{0.947}       & \textbf{0.366}            & \textbf{0.301}       \\ \hline
\multirow{4}{*}{\shortstack[l]{Frozen\\ arm}} & Baseline                     & 0.187            & 0.143       & 0.156            & 0.131            \\
                         & DR & 0.228            & 0.197       & 0.163            & 0.149       \\
                            & RMA                          & 0.318            & 0.178       & 0.191            & 0.148       \\
                            & RALL                 & \textbf{0.987}            & \textbf{0.986}       & \textbf{0.384}            & \textbf{0.317}       \\ \hline
\multirow{4}{*}{\shortstack[l]{Moving\\ arm}} & Baseline                     & 0.299            & 0.246       & 0.259            & 0.206       \\
                            & DR & 0.249            & 0.247       & 0.190            & 0.183       \\
                            & RMA                          & 0.244            & 0.237       & 0.281            & 0.219       \\
                            & RALL                   & \textbf{0.517}           & \textbf{0.437}     & \textbf{0.309}            & \textbf{0.285}       \\
\bottomrule
\end{tabular}
\end{table}

\subsection{Compared Methods}
To quantitatively analyze the improvement in robustness, we conducted experiments with the following methods:
\begin{itemize}
    \item \textbf{Baseline}: The policy is trained using PPO with curriculum terrain setting following \cite{Isaac-Gym-walk}.
    \item \textbf{Expanded Domain Randomization(DR)}: The policy is trained under a broader range of domain randomization compared to \emph{Baseline} to withstand larger disturbances.
    \item \textbf{RMA}: The policy is trained using RMA\cite{RMA-2021}, a method that leverages the teacher-student framework to estimate a latent vector of the environment.
\end{itemize}

The compared methods and RALL are implemented with the same parameter configuration, including PPO parameters, domain randomization ranges (except DR) and observation noise. And the compared methods share the same network configurations with RALL except the value distribution head.

\begin{figure}[t]
    \centering
    \begin{minipage}{0.42\linewidth}
        \centering
        \includegraphics[width=\textwidth]{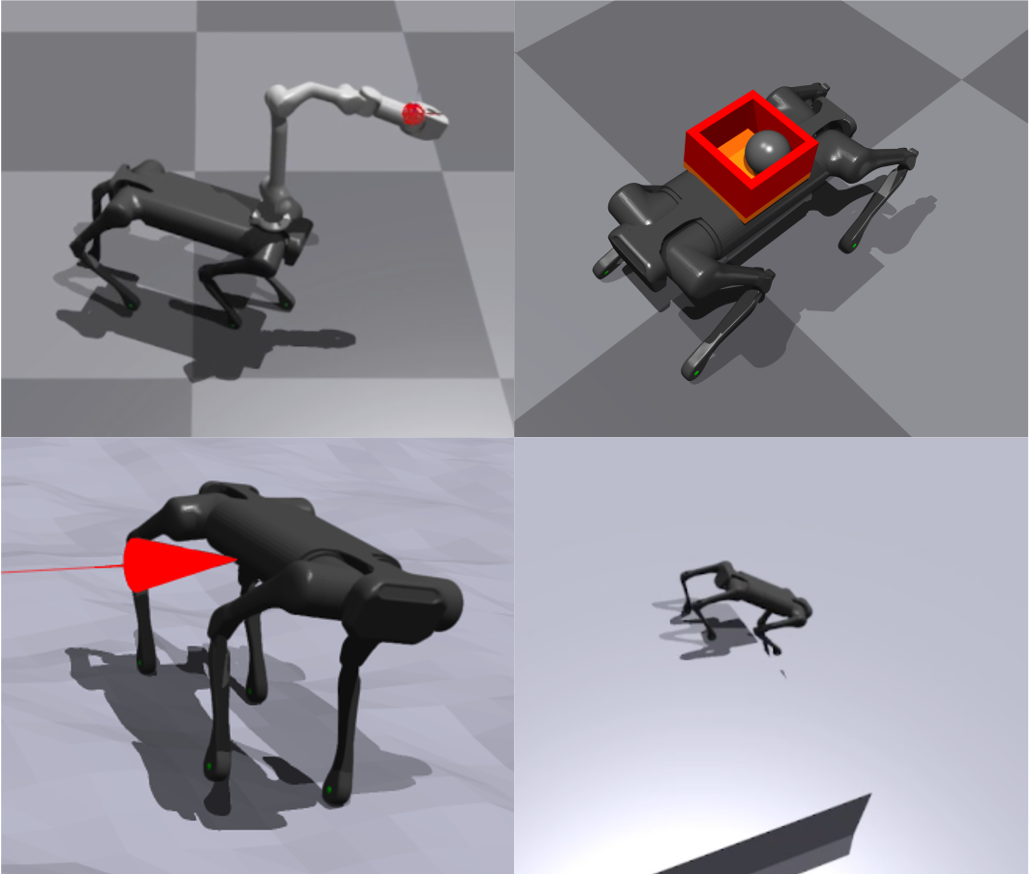}
        \caption{We consider challenging scenarios including dynamic payload, external push and missing a step.}
        \vspace{-10pt}
        \label{fig:sim-set}
    \end{minipage}
    \begin{minipage}{0.55\linewidth}
        \fontsize{10}{12}\selectfont
        \centering
        \captionof{table}{Success rates of robot walking down a platform.}
        \label{tab:result2}
        \begin{tabular}{llc}
        \toprule
        Height  & Policy   & Success \%   \\ \hline
        \multirow{4}{*}{0.4m}  & baseline & 14.4 \\
                              & DR       & 20.6 \\
                              & RMA      & 43.5 \\ 
                              & RALL     & \textbf{92.7} \\ \hline
        \multirow{4}{*}{0.45m} & baseline & 13.9 \\
                              & DR       & 14.3 \\
                              & RMA      & 21.4 \\   
                              & RALL     & \textbf{30.5}\\
        \bottomrule
        \end{tabular}
    \end{minipage}
\end{figure}

\begin{figure*}[t]
    \centering
    \includegraphics[width=\textwidth]{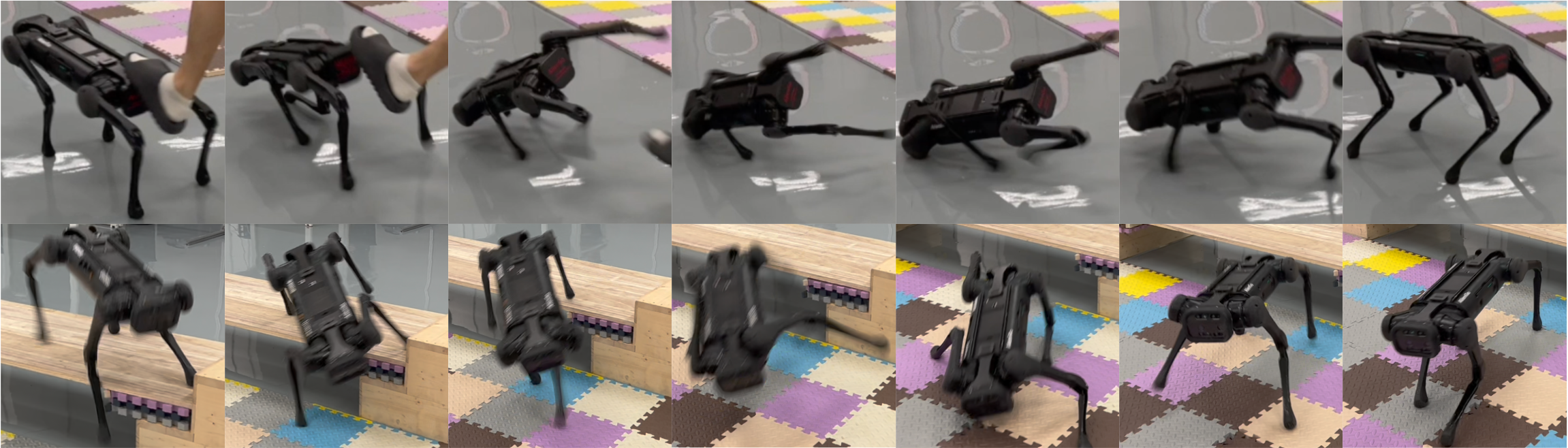}
    \caption{Snapshots of the robot's recovery when it missed a step or got pushed. The robust policy learned through risk-sensitive RL makes the robot retain balance rapidly when encountering intense disturbance.}
    \vspace{-0.4cm}
    \label{fig:real_exp}
\end{figure*}

\begin{figure*}[t]
    \centering
    \includegraphics[width=\textwidth]{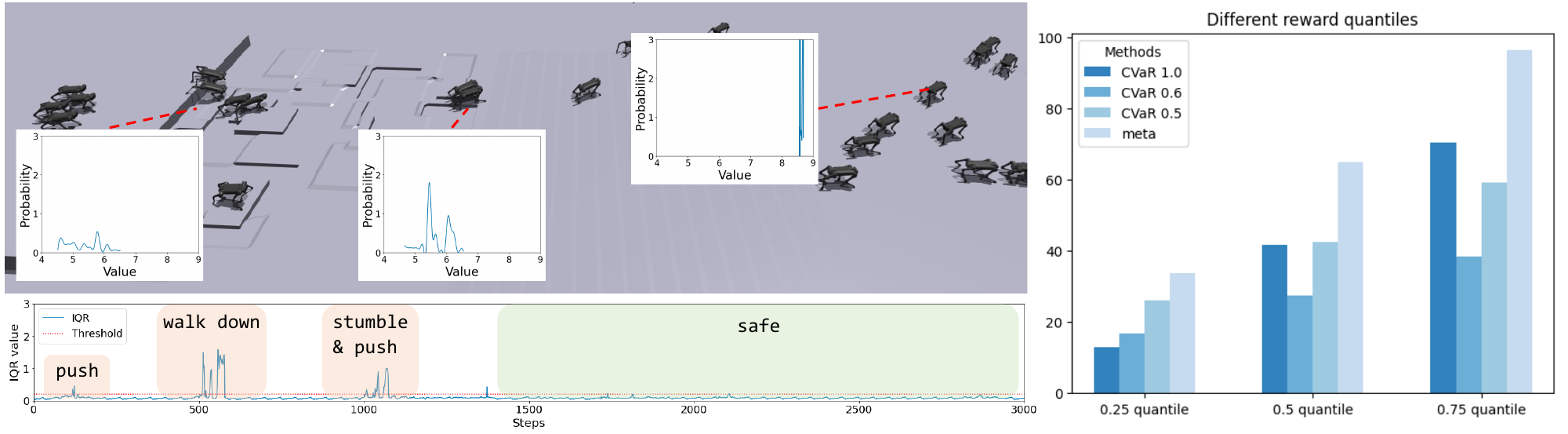}
    \caption{We designed a race track to show the relationship between the IQR value and the current risk level. Robots that are encountering risks exhibit a more dispersed value distribution and a higher IQR value (\textit{left}). We conducted experiments with 50 agents parallelly over 10 episodes. The result shows that the meta controller receives highest return on all quantiles (\textit{right}).}
    \vspace{-0.6cm}
    \label{fig:iqr}
\end{figure*}

\subsection{Results}

To comprehensively assess the robustness of the controller, we consider various factors that could potentially give rise to risks, including loads, external forces, and terrain, as shown in Fig. \ref{fig:sim-set}. Learning methods have been proved effective in enabling robots to walk while carrying loads \cite{bellegarda2022robust}. However, a more challenging scenario arises when the load carried by the robot is dynamic. This will introduce continuous changes to the inertia of the robot, increasing the risk of instability. We conducted three sets of experiments on carrying payloads. The first is carrying a ball that could roll within a rectangular container. The second is attaching a fixed Unitree Z1 robot arm\cite{Z1DexterousRobotic} to Aliengo, and in the third setting we let the arm track random end effector pose when the quadrupedal robot is walking. For each experiment, we introduced two sets of variables: the terrain, which could either be flat or randomly generated rough terrain, and whether to introduce external forces to the robot base. To mitigate randomness, we conducted three trials with different random seeds, and repeated 10 times for each seed. We recorded the \textit{Time to Fall} (TTF) for each experiment, which is the average episode length across all robot instances divided by the total episode length. The results, as shown in Table \ref{tab:result1}, indicate that our method outperforms others by making the robot survive for the longest duration under each setting. 
We also conducted experiments to assess the robot's robustness in handling hazardous terrains in the simulation. We let the robot to walk down a high platform and recorded the success rate of landing. Table \ref{tab:result2} shows that our method has a significantly higher success rate compared to others. 

We conducted numerous experiments on a real Aliengo robot. The results show the effectiveness of our approach in enhancing the robot's robustness. Snapshots of two experiments highlight the robot's ability to regain stability when approaching risks as shown in Fig. \ref{fig:real_exp}. For more real-world experiments, please refer to the supplemental video.

Finally, we conducted experiments on the meta controller proposed in Section \ref{sec:method-c}. We designed a race track that combined both challenging and normal terrains, and randomly applied external forces to the robot throughout the entire episode. Fig. \ref{fig:iqr} showcases the simulation environment and the IQR of the value distribution during the process. It could be seen that when the robot stepped down the platform and traversed the obstacles, the IQR of the value distribution was relatively higher than that of safe scenarios. To validate the effectiveness of the proposed meta controller quantitatively, we conducted experiments in this track with 50 agents parallelly over 10 episodes. The compared policies were obtained by optimizing different CVaR objectives, and the meta controller switched between a $\text{CVaR}_1$ and a $\text{CVaR}_{0.5}$ policy according to the IQR value. We recorded all the returns and calculated the 0.25, 0.5 and 0.75 quantile of the return distribution, which represented the worst-case, middle-case and best-case return, respectively. The result shows that the meta controller achieves the highest return in all worst, middle and best cases, which is consistent with our expectation.

\section{CONCLUSION}
In this work, we present a novel approach to enhancing the robustness of quadrupedal robot locomotion through risk-sensitive reinforcement learning. Experimental results show that our method enables the robot to resist significant disturbances. Moreover, the value distribution given by the critic could serve as an assessment of the current risk level. A limitation of this work is that it only utilizes proprioception to identify risks. For future work, 
we plan to incorporate external sensors to enhance the robot’s capability to estimate environmental risks.



\bibliographystyle{IEEEtran}
\bibliography{reference}

\end{document}